\documentclass[a4, conference]{IEEEtran}

\usepackage{hyperref}
\usepackage{cite}
\usepackage{amsmath,amssymb,amsfonts}
\usepackage{algorithmic}
\usepackage[ruled,vlined]{algorithm2e}
\usepackage{calrsfs}
\usepackage{graphicx}
\usepackage{textcomp}
\usepackage{xcolor}
\usepackage{siunitx}
\usepackage{lipsum}
\usepackage{breqn}
\DeclareMathAlphabet{\pazocal}{OMS}{zplm}{m}{n}

\def\BibTeX{{\rm B\kern-.05em{\sc i\kern-.025em b}\kern-.08em
    T\kern-.1667em\lower.7ex\hbox{E}\kern-.125emX}}
\begin{document}

\title{Building Safer Autonomous Agents by Leveraging Risky Driving Behavior Knowledge}

\author{\IEEEauthorblockN{Ashish Rana}
\IEEEauthorblockA{ Thapar Institute of Eng. \& Tech., India \\ arana_be15@thapar.edu }

\and

\IEEEauthorblockN{Avleen Malhi}
\IEEEauthorblockA{ Bournemouth University, UK \\
amalhi@bournemouth.ac.uk }
}

\maketitle

\begin{abstract}

Simulation environments are good for learning different driving tasks like lane changing, parking or handling intersections etc. in an abstract manner. However, these simulation environments often restrict themselves to operate under conservative interaction behavior amongst different vehicles. But, as we know, real driving tasks often involve very high risk scenarios where other drivers often don't behave in the expected sense. There can be many reasons for this behavior like being tired or inexperienced. The simulation environment doesn't take this information into account while training the navigation agent. Therefore, in this study we especially focus on systematically creating these risk prone scenarios with heavy traffic and unexpected random behavior for creating better model-free learning agents. We generate multiple autonomous driving scenarios by creating new custom Markov Decision Process (MDP) environment iterations in the highway-env simulation package. The behavior policy is learnt by agents trained with the help from deep reinforcement learning models. Our behavior policy is deliberated to handle collisions and risky randomized driver behavior. We train model free learning agents with supplement information of risk prone driving scenarios and compare their performance with baseline agents. Finally, we casually measure the impact of adding these perturbations in the training process to precisely account for the performance improvement obtained from utilizing the learnings from these scenarios.


\end{abstract}

\section{Introduction}

    The arrival of autonomous driving agents have had a great impact on the automobile industry. And it will be responsible for shaping the future of this industry as well. The current industry trend progression demonstrates that the connected fleet of autonomous agents will be dominating our driving infrastructure~\cite{elliott2019recent}. But, that vision is still quite far considering the safety infrastructure, security and public policy reasons~\cite{bagloee2016autonomous, joy2017internet, anderson2014autonomous, litman2020autonomous}. Therefore, our immediate focus should be on making our driving agents safe and efficacious. Creating safer agents has been explored well for the past few years~\cite{li2016intelligence, koren2018adaptive, cui2018development, liu2019road, liu2019safe, thorn2018framework, li2019parallel}. As it is crucial to know the expected agent behavior in different environments especially safety critical ones. In natural driving environments risk prone scenarios don't happen frequently which makes the learning process harder from these scenarios~\cite{yan2021distributionally, feng2021intelligent}. And it is also unethical to create these real risky driving scenarios for experimentation purposes. Therefore, generating and studying these scenarios systematically is a daunting task.

    Perception systems with adversarial approaches of noisy labelling do demonstrate promising path~\cite{ding2020learning, eykholt2018robust, xie2017adversarial}. But the underlying fundamental problem remains focused around safe interactions with other vehicles which themselves are operating independently. Simulation environments with appropriate system dynamics design assumptions do overcome this expressed safety issue. Along with that these environments allow us to systematically study the risk prone scenarios with near realistic driving behaviors. For this study we have used the highway-env simulation package which allows us to simulate different driving tasks~\cite{highway-env}. It also provides simple interfacing capabilities to modify these environments and quickly create baseline prototypes~\cite{rl-agents} on it. We formulate the dynamics of the simulation system as Markov Decision Process (MDP) in our experimentation~\cite{leurent2019approximate}. We model our value function approximation agents with deep reinforcement learning for these defined MDP systems~\cite{mnih2013playing, van2016deep} for our study.

    \begin{figure}
        \centering
        \includegraphics[width=0.455\textwidth]{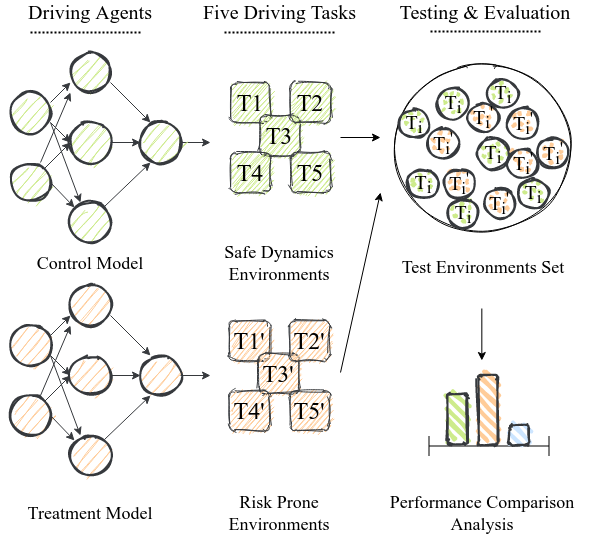}
        \caption{Treatment and Control agents evaluated randomly for safe and perturbed test environments.}
        \label{fig1}
    \end{figure}

    For our experiment design we create two distinct variants of each model architecture for above stated simulation dynamics. One of these model variants is trained with our increased dangerous driving behavior interventions for all the driving related tasks present in highway-env package. Second model variant is the control variable in our experiment which is used to define the reward baseline for the trained agents on regular simulations. In our study we do a methodological doping of these environments with randomized dangerous driving scenarios to create more risk prone environments for driving. This is done specifically in two ways in our study, first by increasing the traffic at strategically important locations challenging the agent to make dangerous overtakes. Second, we increase the randomization factor and the clogged lanes make the environment more collision prone for our agent \textit{(ego-vehicle)}. This is done to create more robust agents in a given environment for that particular dynamic scenario. Which are essentially better at post impact trajectory predictions of other interacting vehicles. Figure~\ref{fig1} explains our causal analysis experimentation setup in sequential form.

\begin{figure}
        \centering
        \includegraphics[width=0.425\textwidth]{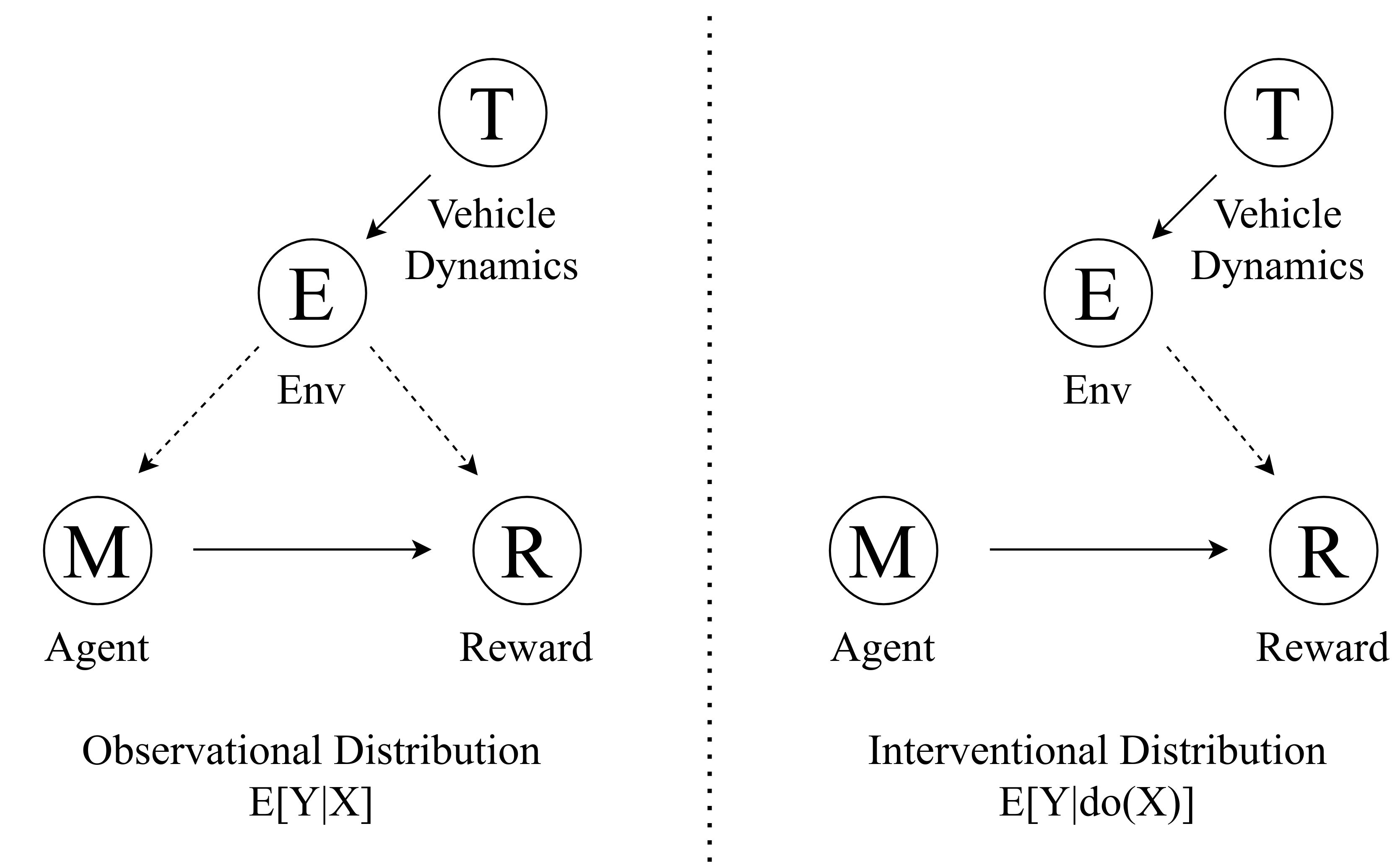}
        \caption{Observational \textit{(Left)} and Interventional \textit{(Right)} distributions represented with causal graphs.}
        \label{fig2}
    \end{figure}

    We also attempt to understand our experimentation setup from a causal standpoint. We hold complete control of the data generation process in our experimentation. Which allows us to have a unique vantage point of conducting an experiment study which is equivalent to a randomized control trial (RCT). We train our agents with a good enough assumption of absence in unobservable confounding variables. As we have strictly defined the state dynamics and vehicle behavior governing models. Our customization capabilities in the highway-env package allows us to keep every condition the same while training except our collision scenario perturbations. Meaning that treatment and control groups are same in all aspects except our treatment i.e. there is a comparability and co-variate balance in our experimental setup~\cite{pearl2009causal, pearl2018book, neuberg2003causality, holland1986statistics}. With this special relation establishment we can imply that association found in our experiment setup is causation. As shown in Figure~\ref{fig2} our treatment \textit{(T)} is subjecting the agent learning process to risk prone interacting vehicle dynamics in a given environment. After that our sample test experiment population involves evaluating the two model variants against regular and perturbed environments with excessive risk prone scenario doping. And finally by using expectation equations derived from the above causal graph we estimate the causal effect our novel learning changes for enhanced safety.

    Our contributions by the means of this paper include providing benchmarking environment simulations for collision robustness predictions. Along with that we provide experimentation methodology that creates more robust agents that provide better on-road safety. And we also causally measure the impact of our risky driving behavior doping interventions for different driving environments. For the remaining paper we first discuss related work corresponding to utilizing risk prone behavior for creating safer autonomous vehicles. Second, we formally define our problem and elaborate on causal aspects of it. Third, we explain the experiment setup for creating these robust agents and elaborate upon our results. Finally, we provide the conclusion to our autonomous driving agent study.

\section{Previous Work}

    Deep Reinforcement Learning has been used extensively for traffic control tasks~\cite{belletti2017expert, wu2017emergent}. The simulated environment provided by CARLA~\cite{dosovitskiy2017carla} gives a framework for systems that estimates several affordances from sensors in a simulated environment~\cite{sauer2018conditional}. Navigation tasks like merging traffic requiring good exploration capabilities have also shown promising results in simulators~\cite{shalev2016safe}. ChauffeurNet~\cite{bansal2018chauffeurnet} elaborates on the idea of imitation learning for training robust autonomous agents that leverages worst case scenarios in the form of realistic perturbations. A clustering based collision case generation study~\cite{sun2021corner} systematically defines and generates the different types of collisions for effectively identifying valuable cases for agent training.

    The highway-env package specifically focuses on designing safe operational policies for large-scale non-linear stochastic autonomous driving systems~\cite{leurent2019approximate}. This environment has been extensively studied and used for modelling different variants of MDP, for example: finite MDP, constraint-MDP and budgeted-MDP (BMDP)~\cite{leurent2020safe}. BMDP is a variant of MDP which makes sure that risk notion implemented as cost signal stays below a certain adjustable threshold~\cite{carrara2019budgeted}. The problem formalization for workings of vehicle kinematics, temporal abstraction, partial observability and reward hypothesis has been studied extensively as well~\cite{leurent2020safe}. Robust optimization planning has been studied in the past for finite MDP systems with uncertain parameters~\cite{ernst2005tree, leurent2019interval, leurent2019practical} and it also has shown promising results under conservative driving behavior. For BMDP, efficacy and safety analysis has been extended into continuous kinematics states and unknown human behavior from the existing known dynamics and finite state space~\cite{carrara2019budgeted}. Model free learning networks that approximate value function for these MDPs like Deep Q-Learning (DQN) and Dueling Deep Q-Learning (DQN) networks have demonstrated promising results in continuous agent learning~\cite{gu2016continuous, wang2016dueling}.

    Worst case scenario knowledge in traffic analysis has been leveraged for model based algorithms by building regions of high confidence containing true dynamics with high probability. Tree based planning algorithms were used to achieve robust stabilisation and mini-max control with generic costs. These studies also leveraged non-asymptotic linear regression and interval prediction~\cite{leurent2020safe, leurent2019approximate} for safer trajectory predictions. Behavior guided action studies which use proximity graphs and safety trajectory computes for working with aggressive and conservative have shown promising results~\cite{mavrogiannis2020b} as well. This study used CMetric measure~\cite{chandra2020cmetric} for generating varying levels of aggressiveness in traffic. Whereas in our case we have used more randomization and traffic clogging at key areas for risk prone scenarios to measure more granular observation results.

    Causal modelling techniques have contributed a lot in terms of providing great interpretative explanations in many domains~\cite{hicks1980causality, smirnov2009granger, granger1988some, pearl2018does}. Also, Fischer's randomized control trials (RCTs) have served as the gold standard for causal discovery from observational data~\cite{pearl2018book}. Sewall's path diagrams were the first attempt of generating causal answers with mathematics~\cite{wright1934method}. Now, causal diagrams and different adjustments on these diagrams do offer direct causal relation information about any experimental variables under study~\cite{pearl2009causal, pearl2018book, richardson2013single, bareinboim2020pearl}. In our experimental study we use these existing mathematical tools to draw direct causal conclusions of our learnings from our environment interventions.

\section{Problem Formulation}

    Our goal is to design and build agents on collision prone MDPs for navigation tasks across the different traffic scenarios. The MDP comprises a behavior policy $\pi$(\textit{a} $\mid$ \textit{s}) that outputs action \textit{a} for a given state \textit{s}. With this learnt policy our goal is to predict discrete safe and efficient action from finite action set \textit{(left, right, break, accelerate, idle)} for next time step for given driving scenarios.

    The simulation platform that we used is compatible with OpenAI gym package~\cite{brockman2016openai}. The highway-env package provides the traffic flow which is governed by Intelligent Driver Model (IDM)~\cite{treiber2000congested} for linear acceleration \& MOBIL model~\cite{kesting2007general} for lane changing. MOBIL model primarily consists of safety criterion and incentive criterion. First safety criterion checks whether after lane change the given vehicle is having enough acceleration space and second criterion determines the total advantage of lane change in terms of total acceleration gain.

    Given MDP \( \pazocal{M} \) is defined as a set of \textit{(\( \pazocal{S} \), \( \pazocal{A} \), \( \pazocal{T} \), r)} where action \textit{a} $\in$ \( \pazocal{A} \), state \textit{s} $\in$ \( \pazocal{S} \), reward function \textit{r} $\in$ [0,1]\textsuperscript{ \( \pazocal{S} \) X \( \pazocal{A} \) } and state transition probabilities T\textit{(s' $\mid$ s, a)} $\in$ \textit{\( \pazocal{M} \)(\( \pazocal{S} \))\textsuperscript{ \( \pazocal{S} \) X \( \pazocal{A} \) }}. With Deep-RL algorithms we search the the behavior policy $\pi$(s$\mid$a) that helps us in navigating across the traffic environments to gather maximum discounted reward \( \pazocal{R} \). The state-action value function \textit{\( \pazocal{Q} \)\textsuperscript{$\pi$}(s,a) } for given \textit{(s,a)} assists in estimating future rewards of given behavior policy $\pi$. Therefore, the optimal state-action value function \textit{\( \pazocal{Q} \)\textsuperscript{$*$}(s,a)} provides maximum value estimates for all \textit{s} $\in$ \( \pazocal{S} \) and is evaluated by solving Bellman Equation~\cite{bellman2015applied}, stated below for reference. From this the optimal policy $\pi$ is expressed as \textit{\textbf{$\pi$(s)}} = arg max\textsubscript{a$\in$\(\pazocal{A}\)}\textit{\( \pazocal{Q} \)\textsuperscript{$*$}(s,a)}. We used DQN with duelling network architecture having \textit{value \& advantage streams} to approximate the state-action value function \textit{ \( \pazocal{Q} \)(s,a) } which predicts best possible action as learned from the policy $\pi$.

    \begin{equation}
        Q \textsuperscript{*} (s, a) = \mathop{\mathbb{E}} [ R(s,a) + \gamma \sum \limits _{s'} P (s' | s , a) \max \limits _{a'} Q \textsuperscript{*} (s', a') ]
    \end{equation}

    From our experimentation setup we intend to derive direct causal impact of our interventions in traffic environment scenarios. And as we can refer back from Figure~\ref{fig2}, our treatment (T) is subjecting the learning agent to more risky driving behavior. Our testing sample set involves random agent reward calculation against perturbed and control environments which makes our experiment equivalent to RCT. Meaning that there is no unobserved confounding present in our experimentation i.e. backdoor criterion is satisfied. Also, in RCTs distribution of all co-variates are same except the treatment. Co-variate balance in observational data also implies that association is equal to causation while calculating the potential outcomes, refer equation stated below.

    \begin{equation}
        \begin{alignedat}{3}
            P(X \mid T=1) \stackrel{d}{=} P(X \mid T=0) \\
            P(X \mid T=1) \stackrel{d}{=} P(X),  T \perp\!\!\!\perp X \\
            P(X \mid T=0) \stackrel{d}{=} P(X), T \perp\!\!\!\perp X
        \end{alignedat}
    \end{equation}

    Essentially meaning that we can use the associative difference quantity to infer the effect of treatment on outcomes. Meaning that we can use the Average Treatment Effect (ATE) approach for calculating the causal effect by simply subtracting the averaged out values treatment and control potential outcomes. In below stated equations Y(1) $\triangleq$ Y\textsubscript{ i $\mid$ do(T=1)} \& Y(0) $\triangleq$ Y\textsubscript{ i $\mid$ do(T=0)} and these equations hold true in case of RCTs where causal difference can be calculated with associated difference.

    \begin{equation}
        \begin{alignedat}{2}\label{ate}
            E[Y(1)-Y(0)] &= E[Y(1)]-E[Y(0)] \\
            E[Y(1)]-E[Y(0)] &= E[Y\mid T=1]-E[Y\mid T=0] 
        \end{alignedat}
    \end{equation}

    \begin{figure*}
        \centering
        \includegraphics[width=0.965\textwidth]{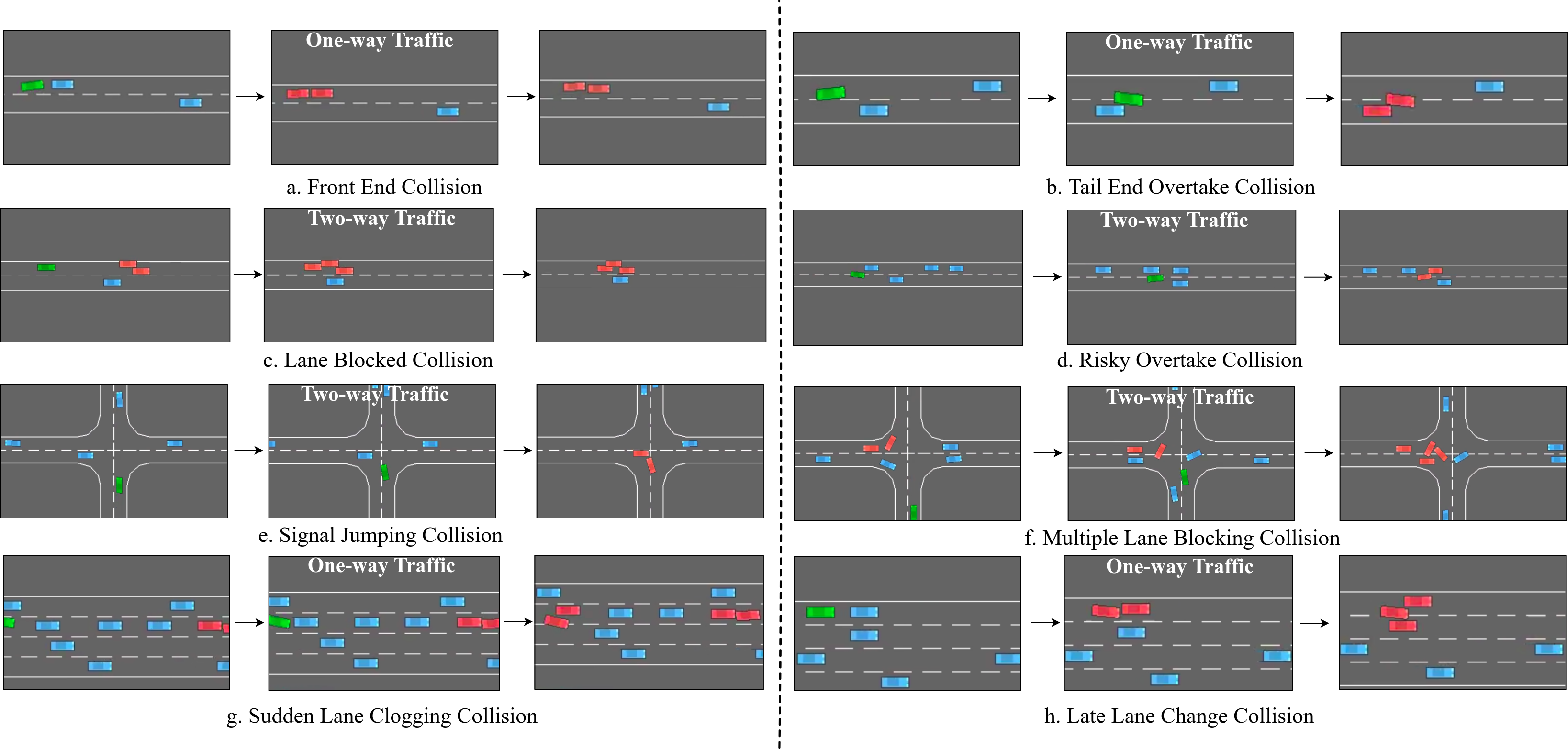}
        \caption{Possibly realistic collision examples simulated from the different driving tasks in highway-env package environments.}
        \label{fig3}
    \end{figure*}

    Model free learning approaches generally don't have explicit information about the dynamics of the systems. Therefore, during the training these agents will generalize their policies corresponding to the particular given scenarios only. We introduce risky randomized behavior in these environment vehicles with collision prone randomized behavior. Figure~\ref{fig3} shows different probable real life collisions simulated in different highway-env package environments. It increases generalizations on less common but highly critical scenarios which can save users from hefty collisions. We critically analyze the performance of our treated agents in comparison to control agents for important tasks like performing roundabouts, handling intersections, u-turn \& two-way traffic overtakes and lane changing environments in our study.

    \begin{figure}
        \centering
        \includegraphics[width=0.40\textwidth]{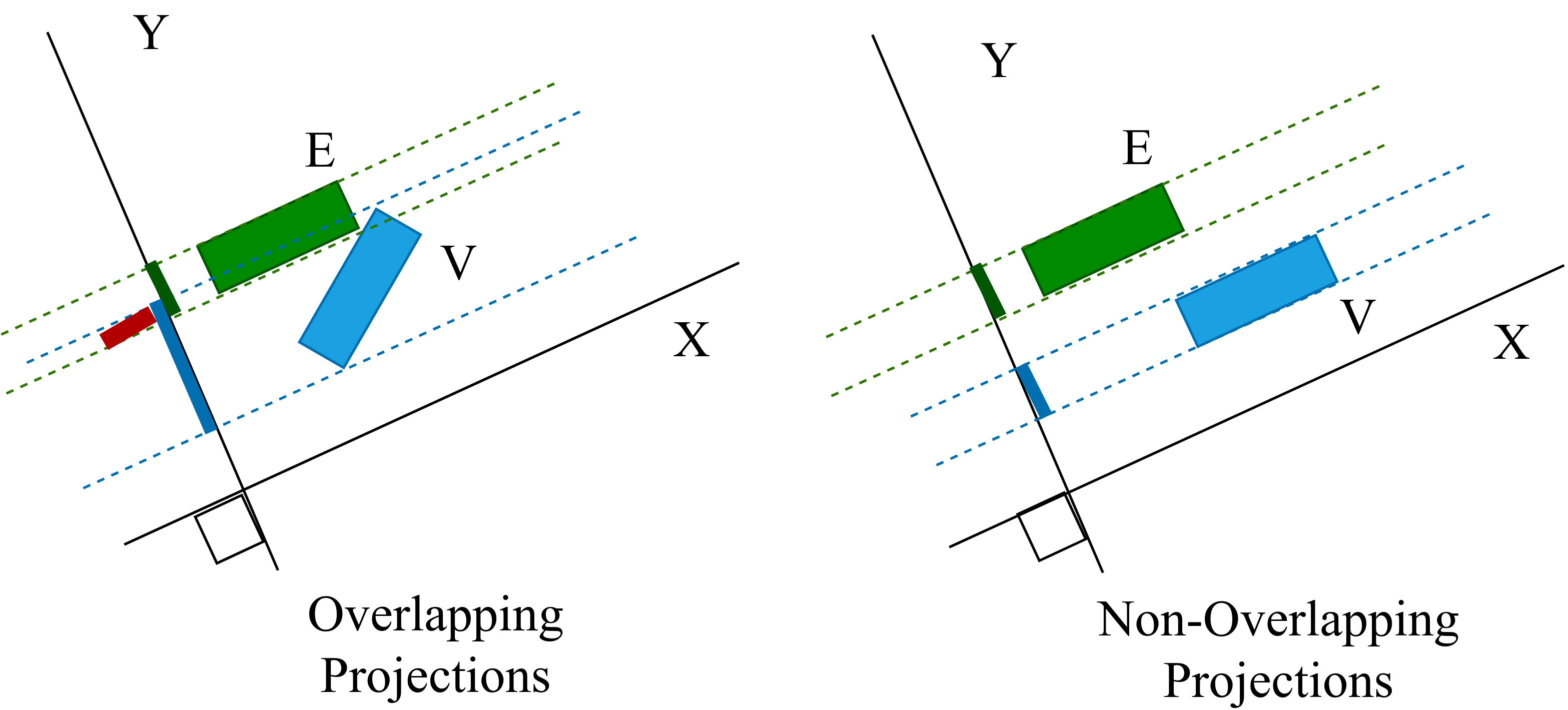}
        \caption{Overlapping \textit{(Left)} and Non-Overlapping \textit{(Right)} projection analysis for collision detection in highway-env package.}
        \label{fig4}
    \end{figure}

    The collision between two vehicles is equivalent to intersection of two polygons in the rendered environment output of highway-env. And we detect these collisions between rectangular polygons with the separating axis theorem for the given two convex polygons. Essentially, the idea is to find a line that separates both polygons. If that line exists then polygons are separated and collision hasn't happened yet. Algorithmically for each edge of our base rectangular polygon we find a perpendicular axis to current edges under review. After that we project these edges onto that axis and in case these projections don't overlap it functionally means no collision as rectangular polygons are not intersecting, refer Figure~\ref{fig4}.

\section{Experiment Setup}

\begin{figure}
        \centering
            \includegraphics[width=0.46\textwidth]{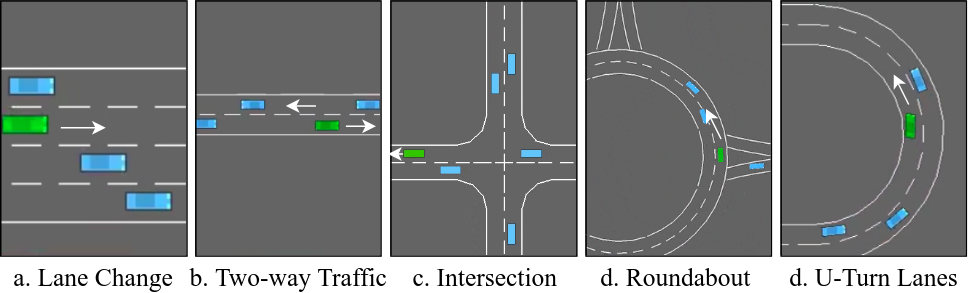}
            \caption{Different driving related navigation environment tasks for collision prone scenario analysis.}
        \label{fig5}
    \end{figure}

    In our experimentation setup we calculate the ATE metric for namely lane changing, two-way traffic, roundabout, intersection and u-turn tasks, refer Figure~\ref{fig5}. These five tasks are evaluated against increasing traffic from default vehicle count/density to a 200\% increase. For each of these traffic scenarios we create our treatment environment with varying degrees of acceleration \& steering parameters which would not comply with MOBIL model criteria of safety and incentives. This randomization behavior is governed by equations stated below which lays the foundation of simulating collision prone behavior. We create collision prone behavior by significantly changing the equation max-min acceleration parameters in \textit{create\_random()} function defined in the kinematics rules of the highway-env package. With our experimentation setup we quantify the causal model performance improvements from introduction of this risk factor knowledge in the agent learning process. And compare it with the control baseline models across these five different navigation tasks against a spectrum of increasing traffic density.

    \begin{equation}
        \begin{alignedat}{2}\label{acc-eq}
            acc_p = acc\textsubscript{min} + \textit{rand[0,1]}*(acc\textsubscript{max} - acc\textsubscript{min}) \\
            str_p = str\textsubscript{min} + \textit{rand[0,1]}*(str\textsubscript{max} - str\textsubscript{min})
        \end{alignedat}
    \end{equation}

    We rebuild the highway-env package environments with our custom changes by altering the environment configurations, randomizing behavior and adding new vehicles to strategically increase the traffic density in order to simulate risky driver behavior. In the lane changing task for treatment \& control model training we incrementally increase the vehicle count from 50 to 150 vehicles accompanied with an equivalent intermittent increase of vehicle density by 100\% and having increased randomized risky behavior on episode duration length of 20 seconds. We train unique agents for navigation on each different traffic count environments in our experimentation setup corresponding to every vehicle count increase. Also, we plot a comparative analysis performance graph of control and treatment agent across these different environment iterations and calculate ATE of our perturbations. Similarly, for the u-turn environment we uniformly increase our vehicle count from 3 to 12 with incremental increase of vehicle count of 3. Also, for two-way traffic we reduce original environment length to 2/3\textsubscript{rd} of the original and incrementally increase the vehicle traffic count from base 5 in direction and 2 in opposite direction vehicles to 15 in direction and 6 in opposite direction vehicles. For collision prone treatment in intersection task driving tasks we rewire the randomization behavior to a more risky one with our acceleration and deceleration tuning. We also increase the vehicle count from 10 to 30 incrementally with an interval gap of 5 vehicles and alongside we incrementally increase the spawning probability by 0.1 until it reaches its maximum value. Finally, for roundabout tasks we incrementally increase the traffic from 5 to 15 vehicles with risk prone randomization in our treatment environment for agent training and performance comparison with control baseline. Each of these environment configurations requires the environment to be rebuilt iteratively. Additionally, repeated model training against each new treatment \& control environment samples with different initiating random seeds. For evaluating our model's performance we have kept the traffic as constant in our test population set against the corresponding treatment and control environment on which agents were trained. And we have only changed the risky behavior in treatment and control environment sets to calculate the ATE for measuring causal performance improvement.

    \begin{figure}
        \centering
        \includegraphics[width=0.4\textwidth]{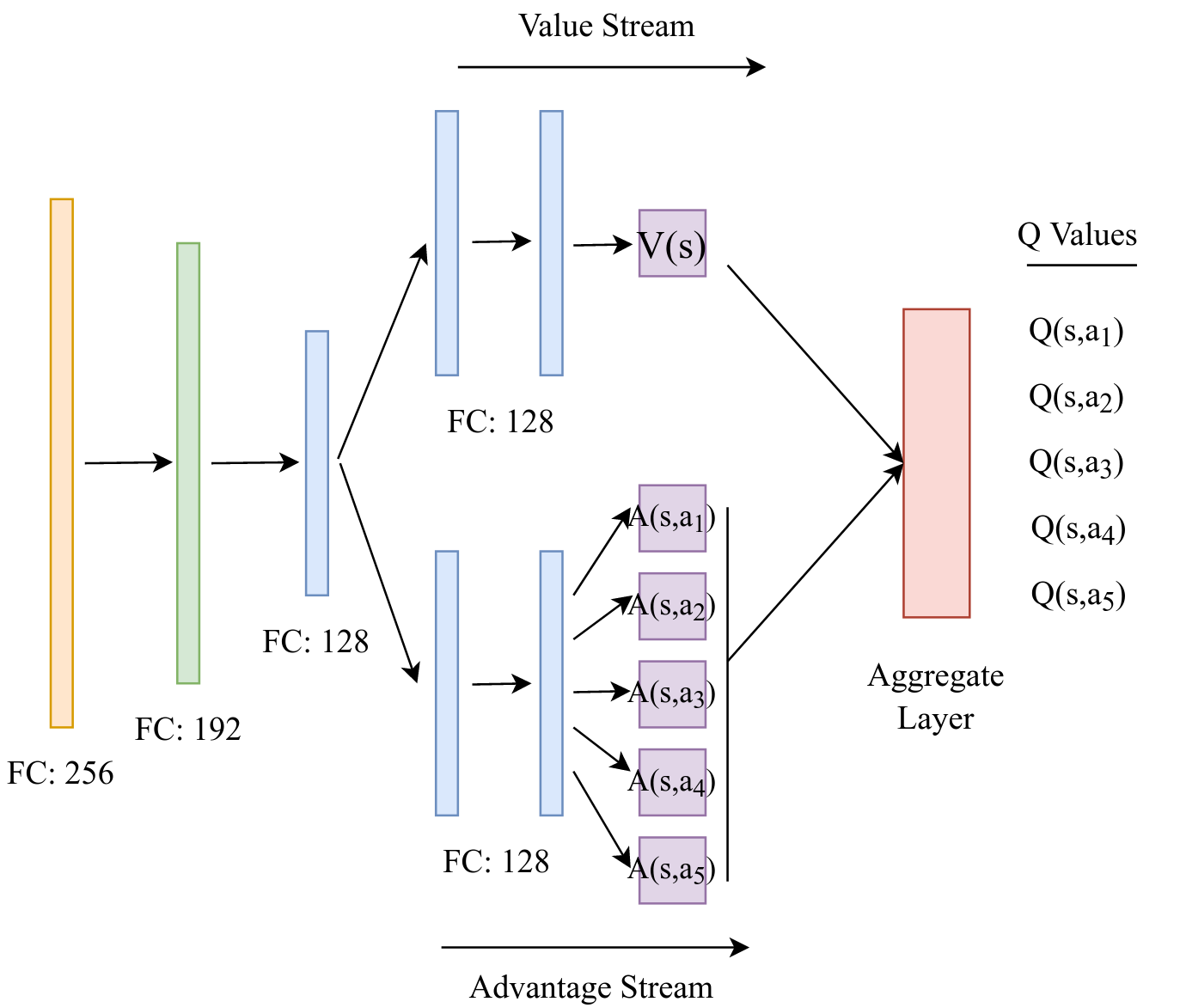}
        \caption{Dueling Deep Q-Learning baseline model architecture used for agent training.}
        \label{fig6}
    \end{figure}

    We use the DQN reinforcement learning modelling technique with ADAM optimizer for our experiment with a learning rate of 5e-4. Our discount factor used is 0.99 and environment observation vector data is fed in a batch size of 100. Our agents are trained over 3072 episodes until they converge to average reward from a given driving environment. We use the dueling network design which utilizes advantage function \textit{A(s,a)} which helps in estimating state-action value function \textit{Q(s,a)} for state-action pairs more precisely. This is done by splitting the network into two streams, value and advantage ones which share some base hidden layers. The shared network consists of 3 fully connected layers of 256, 192 and 128 units respectively. Value and advantage stream consists of 2 layers of 128 units each. The final output of these streams is also fully connected to the network. The value stream has 1 output of the calculated value function for a given state. And the advantage stream has \textit{n\textsubscript{a}} outputs representing the number of discrete possible actions for a given state. The output vectors from these two streams are combined to calculate the state-action value function estimate with the help from below stated equation, refer Figure~\ref{fig6} for model architecture. \footnote{Baseline DQN agent implementations referenced from: github.com/eleurent/rl-agents}

    \begin{equation}
        Q(s, a) = V (s) + A(s, a)
    \end{equation}

\section{Results}

    Another randomization factor in our experimentation setup involves initial random seed values.This is done for adding random behavior to tasks like randomizing vehicle acceleration \& deceleration, spawning vehicles, vehicle location etc. Therefore, we test our trained treatment and control models against the risk-prone and regular driving environments with different randomization seed values to average any anomalous results. Hence, we measure the ATE by evaluating our agents against several randomized seed values for risk prone and regular driving environments. This calculation is simply summarized by the equation below where summation of first two terms evaluate the average reward calculated from treatment models in risk prone and regular environment samples. The last two terms calculate the same for control agent in both these environment samples again respectively and the test set sample count is expressed as \( \pazocal{N} \) \textsubscript{T} = \( \pazocal{N} \) \textsubscript{C} = 100. The associated difference of these quantities gives us the ATE for performance improvements in our robust agents trained on perturbed environments as explained earlier in the problem formulation section.

    \begin{dmath}
        ATE=1/N\textsubscript{T, C}*[\sum_{i=1}^{\textsubscript{N\textsubscript{T}}}R^T\textsubscript{avg\textsuperscript{i}}+\sum_{i=1}^{\textsubscript{N\textsubscript{C}}}R^T\textsubscript{avg\textsuperscript{i}}-\sum_{i=1}^{\textsubscript{N\textsubscript{T}}}R^C\textsubscript{avg\textsuperscript{i}}+\sum_{i=1}^{\textsubscript{N\textsubscript{C}}} R^C\textsubscript{avg\textsuperscript{i}} ]\
    \end{dmath}

    \begin{figure}
        \centering
        \includegraphics[width=0.35\textwidth]{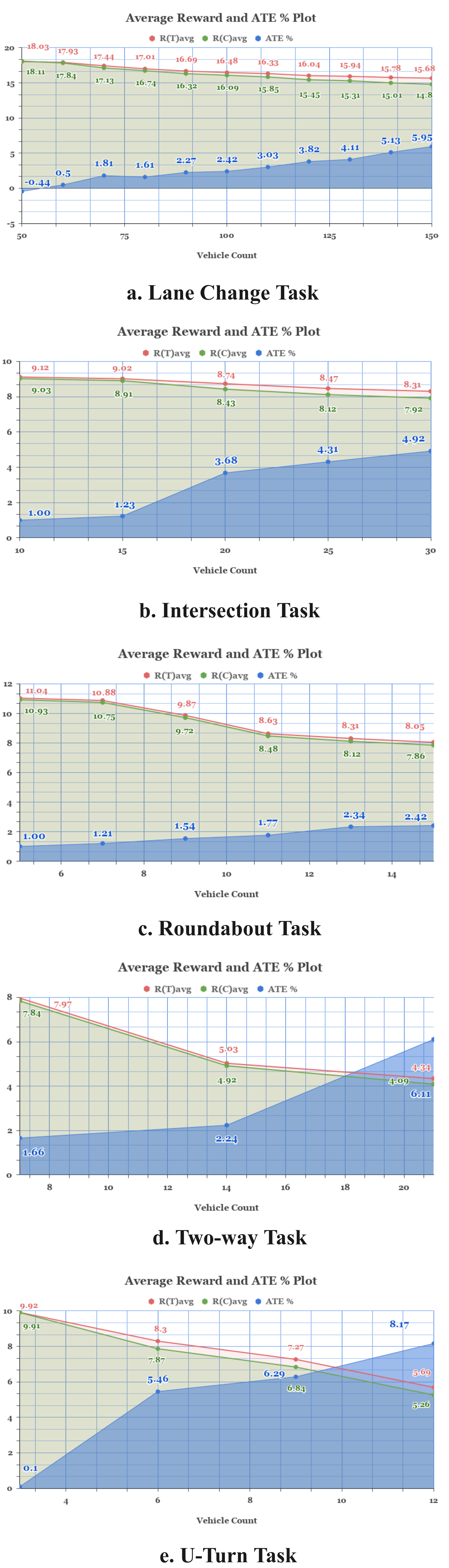}
        \caption{ATE plots for quantifying the performance improvements attained by leveraging the knowledge learnt from collision prone driving scenarios.}
        \label{fig7}
    \end{figure}

    ATE results from Figure~\ref{fig7} clearly demonstrates to us the advantage of teaching agents these risk prone critical scenarios. \footnote{For better readability purposes we converted the calculated ATE values to their respective percentages in Figure~\ref{fig7}.} Across all tasks we observe that as traffic density increases like real-life scenarios in heavily populated cities the positive effect of knowing perturbed scenarios is pronounced for our robust treatment agents. More importantly there is also a declining trend of average reward as the traffic increases for every driving task analyzed for highway-env. This depreciation in agent performance both for control and treatment models can be attributed to constantly decreasing safety distance amongst all vehicles causing more than expected collisions and slow progression by our ego-vehicle across these environments due to heavy traffic. Even with decreasing average reward trend across our test set environment samples the performance of treatment models has always exceeded the control models. Also, the relative improvements in ATE values further increases as the traffic continues to increase demonstrating strong robustness of treatment agents.

    Currently our scope of work is limited to few but critically important driving scenarios. Also, we have used only homogeneous agents in our analysis and attempted to analyze critical knowledge leveraging components on single agent only i.e. our ego-vehicle. Plus our randomization mechanism though uniform doesn't necessarily follow human-like behavior while generating risk prone scenarios. But, our causal effect estimation approach that quantifies the information learnt from perturbed scenarios does demonstrate promising results. It holds a vast scope of practical applications for creating more interpretable, metric-oriented \& key performance indicator \textit{ (KPI)} driven autonomous agent systems.

\section{Conclusion}

    Our experiments from this paper provide insights into the importance of using deliberate interventions of collision prone behaviors while training agents for stochastic processes like autonomous driving. By using the MDP formulation of the discussed driving scenarios with collision simulation perturbations we were able to generate more robust agents. Our treatment model experimentation setup used episode data from traffic clogged lanes and risky randomized behavior during training which finally resulted in positive ATE result values. Which proved that agents trained with a wider range of collision prone scenarios perform better than the existing vanilla simulation agents. Also, we casually quantified the impact of our interventions for the discussed model free learning DQN technique which assisted us in accurately estimating the performance improvements. For every driving scenario environment our new agents produced better results and were proved to be better collision deterrents. Therefore, underscoring the importance of learning valuable lessons from risk prone scenario simulations for creating safe autonomous driving agents.

\bibliographystyle{IEEEtran}

\bibliography{main.bib}

\begin{thebibliography}{10}
\providecommand{\url}[1]{#1}
\csname url@samestyle\endcsname
\providecommand{\newblock}{\relax}
\providecommand{\bibinfo}[2]{#2}
\providecommand{\BIBentrySTDinterwordspacing}{\spaceskip=0pt\relax}
\providecommand{\BIBentryALTinterwordstretchfactor}{4}
\providecommand{\BIBentryALTinterwordspacing}{\spaceskip=\fontdimen2\font plus
\BIBentryALTinterwordstretchfactor\fontdimen3\font minus
  \fontdimen4\font\relax}
\providecommand{\BIBforeignlanguage}[2]{{%
\expandafter\ifx\csname l@#1\endcsname\relax
\typeout{** WARNING: IEEEtran.bst: No hyphenation pattern has been}%
\typeout{** loaded for the language `#1'. Using the pattern for}%
\typeout{** the default language instead.}%
\else
\language=\csname l@#1\endcsname
\fi
#2}}
\providecommand{\BIBdecl}{\relax}
\BIBdecl

\bibitem{elliott2019recent}
D.~Elliott, W.~Keen, and L.~Miao, ``Recent advances in connected and automated
  vehicles,'' \emph{journal of traffic and transportation engineering (English
  edition)}, vol.~6, no.~2, pp. 109--131, 2019.

\bibitem{bagloee2016autonomous}
S.~A. Bagloee, M.~Tavana, M.~Asadi, and T.~Oliver, ``Autonomous vehicles:
  challenges, opportunities, and future implications for transportation
  policies,'' \emph{Journal of modern transportation}, vol.~24, no.~4, pp.
  284--303, 2016.

\bibitem{joy2017internet}
J.~Joy and M.~Gerla, ``Internet of vehicles and autonomous connected
  car-privacy and security issues,'' in \emph{2017 26th International
  Conference on Computer Communication and Networks (ICCCN)}.\hskip 1em plus
  0.5em minus 0.4em\relax IEEE, 2017, pp. 1--9.

\bibitem{anderson2014autonomous}
J.~M. Anderson, K.~Nidhi, K.~D. Stanley, P.~Sorensen, C.~Samaras, and O.~A.
  Oluwatola, \emph{Autonomous vehicle technology: A guide for
  policymakers}.\hskip 1em plus 0.5em minus 0.4em\relax Rand Corporation, 2014.

\bibitem{litman2020autonomous}
T.~Litman, ``Autonomous vehicle implementation predictions: Implications for
  transport planning,'' 2020.

\bibitem{li2016intelligence}
L.~Li, W.-L. Huang, Y.~Liu, N.-N. Zheng, and F.-Y. Wang, ``Intelligence testing
  for autonomous vehicles: A new approach,'' \emph{IEEE Transactions on
  Intelligent Vehicles}, vol.~1, no.~2, pp. 158--166, 2016.

\bibitem{koren2018adaptive}
M.~Koren, S.~Alsaif, R.~Lee, and M.~J. Kochenderfer, ``Adaptive stress testing
  for autonomous vehicles,'' in \emph{2018 IEEE Intelligent Vehicles Symposium
  (IV)}.\hskip 1em plus 0.5em minus 0.4em\relax IEEE, 2018, pp. 1--7.

\bibitem{cui2018development}
L.~Cui, J.~Hu, B.~B. Park, and P.~Bujanovic, ``Development of a simulation
  platform for safety impact analysis considering vehicle dynamics, sensor
  errors, and communication latencies: Assessing cooperative adaptive cruise
  control under cyber attack,'' \emph{Transportation research part C: emerging
  technologies}, vol.~97, pp. 1--22, 2018.

\bibitem{liu2019road}
P.~Liu, Z.~Xu, and X.~Zhao, ``Road tests of self-driving vehicles: affective
  and cognitive pathways in acceptance formation,'' \emph{Transportation
  research part A: policy and practice}, vol. 124, pp. 354--369, 2019.

\bibitem{liu2019safe}
P.~Liu, R.~Yang, and Z.~Xu, ``How safe is safe enough for self-driving
  vehicles?'' \emph{Risk analysis}, vol.~39, no.~2, pp. 315--325, 2019.

\bibitem{thorn2018framework}
E.~Thorn, S.~C. Kimmel, M.~Chaka, B.~A. Hamilton \emph{et~al.}, ``A framework
  for automated driving system testable cases and scenarios,'' United States.
  Department of Transportation. National Highway Traffic Safety~…, Tech.
  Rep., 2018.

\bibitem{li2019parallel}
L.~Li, X.~Wang, K.~Wang, Y.~Lin, J.~Xin, L.~Chen, L.~Xu, B.~Tian, Y.~Ai,
  J.~Wang \emph{et~al.}, ``Parallel testing of vehicle intelligence via
  virtual-real interaction,'' \emph{Science robotics}, vol.~4, no.~28, p.
  eaaw4106, 2019.

\bibitem{yan2021distributionally}
X.~Yan, S.~Feng, H.~Sun, and H.~X. Liu, ``Distributionally consistent
  simulation of naturalistic driving environment for autonomous vehicle
  testing,'' \emph{arXiv preprint arXiv:2101.02828}, 2021.

\bibitem{feng2021intelligent}
S.~Feng, X.~Yan, H.~Sun, Y.~Feng, and H.~X. Liu, ``Intelligent driving
  intelligence test for autonomous vehicles with naturalistic and adversarial
  environment,'' \emph{Nature communications}, vol.~12, no.~1, pp. 1--14, 2021.

\bibitem{ding2020learning}
W.~Ding, B.~Chen, M.~Xu, and D.~Zhao, ``Learning to collide: An adaptive
  safety-critical scenarios generating method,'' in \emph{2020 IEEE/RSJ
  International Conference on Intelligent Robots and Systems (IROS)}.\hskip 1em
  plus 0.5em minus 0.4em\relax IEEE, 2020, pp. 2243--2250.

\bibitem{eykholt2018robust}
K.~Eykholt, I.~Evtimov, E.~Fernandes, B.~Li, A.~Rahmati, C.~Xiao, A.~Prakash,
  T.~Kohno, and D.~Song, ``Robust physical-world attacks on deep learning
  visual classification,'' in \emph{Proceedings of the IEEE Conference on
  Computer Vision and Pattern Recognition}, 2018, pp. 1625--1634.

\bibitem{xie2017adversarial}
C.~Xie, J.~Wang, Z.~Zhang, Y.~Zhou, L.~Xie, and A.~Yuille, ``Adversarial
  examples for semantic segmentation and object detection,'' in
  \emph{Proceedings of the IEEE International Conference on Computer Vision},
  2017, pp. 1369--1378.

\bibitem{highway-env}
E.~Leurent, ``An environment for autonomous driving decision-making,''
  \url{https://github.com/eleurent/highway-env}, 2018.

\bibitem{rl-agents}
------, ``rl-agents: Implementations of reinforcement learning algorithms,''
  \url{https://github.com/eleurent/rl-agents}, 2018.

\bibitem{leurent2019approximate}
E.~Leurent, Y.~Blanco, D.~Efimov, and O.-A. Maillard, ``Approximate robust
  control of uncertain dynamical systems,'' \emph{arXiv preprint
  arXiv:1903.00220}, 2019.

\bibitem{mnih2013playing}
V.~Mnih, K.~Kavukcuoglu, D.~Silver, A.~Graves, I.~Antonoglou, D.~Wierstra, and
  M.~Riedmiller, ``Playing atari with deep reinforcement learning,''
  \emph{arXiv preprint arXiv:1312.5602}, 2013.

\bibitem{van2016deep}
H.~Van~Hasselt, A.~Guez, and D.~Silver, ``Deep reinforcement learning with
  double q-learning,'' in \emph{Proceedings of the AAAI Conference on
  Artificial Intelligence}, vol.~30, no.~1, 2016.

\bibitem{pearl2009causal}
J.~Pearl \emph{et~al.}, ``Causal inference in statistics: An overview,''
  \emph{Statistics surveys}, vol.~3, pp. 96--146, 2009.

\bibitem{pearl2018book}
J.~Pearl and D.~Mackenzie, \emph{The book of why: the new science of cause and
  effect}.\hskip 1em plus 0.5em minus 0.4em\relax Basic books, 2018.

\bibitem{neuberg2003causality}
L.~G. Neuberg, ``Causality: Models, reasoning, and inference,'' 2003.

\bibitem{holland1986statistics}
P.~W. Holland, ``Statistics and causal inference,'' \emph{Journal of the
  American statistical Association}, vol.~81, no. 396, pp. 945--960, 1986.

\bibitem{belletti2017expert}
F.~Belletti, D.~Haziza, G.~Gomes, and A.~M. Bayen, ``Expert level control of
  ramp metering based on multi-task deep reinforcement learning,'' \emph{IEEE
  Transactions on Intelligent Transportation Systems}, vol.~19, no.~4, pp.
  1198--1207, 2017.

\bibitem{wu2017emergent}
C.~Wu, A.~Kreidieh, E.~Vinitsky, and A.~M. Bayen, ``Emergent behaviors in
  mixed-autonomy traffic,'' in \emph{Conference on Robot Learning}.\hskip 1em
  plus 0.5em minus 0.4em\relax PMLR, 2017, pp. 398--407.

\bibitem{dosovitskiy2017carla}
A.~Dosovitskiy, G.~Ros, F.~Codevilla, A.~Lopez, and V.~Koltun, ``Carla: An open
  urban driving simulator,'' in \emph{Conference on robot learning}.\hskip 1em
  plus 0.5em minus 0.4em\relax PMLR, 2017, pp. 1--16.

\bibitem{sauer2018conditional}
A.~Sauer, N.~Savinov, and A.~Geiger, ``Conditional affordance learning for
  driving in urban environments,'' in \emph{Conference on Robot
  Learning}.\hskip 1em plus 0.5em minus 0.4em\relax PMLR, 2018, pp. 237--252.

\bibitem{shalev2016safe}
S.~Shalev-Shwartz, S.~Shammah, and A.~Shashua, ``Safe, multi-agent,
  reinforcement learning for autonomous driving,'' \emph{arXiv preprint
  arXiv:1610.03295}, 2016.

\bibitem{bansal2018chauffeurnet}
M.~Bansal, A.~Krizhevsky, and A.~Ogale, ``Chauffeurnet: Learning to drive by
  imitating the best and synthesizing the worst,'' \emph{arXiv preprint
  arXiv:1812.03079}, 2018.

\bibitem{sun2021corner}
H.~Sun, S.~Feng, X.~Yan, and H.~X. Liu, ``Corner case generation and analysis
  for safety assessment of autonomous vehicles,'' \emph{arXiv preprint
  arXiv:2102.03483}, 2021.

\bibitem{leurent2020safe}
E.~Leurent, ``Safe and efficient reinforcement learning for behavioural
  planning in autonomous driving,'' Ph.D. dissertation, Universit{\'e} de
  Lille, 2020.

\bibitem{carrara2019budgeted}
N.~Carrara, E.~Leurent, R.~Laroche, T.~Urvoy, O.-A. Maillard, and O.~Pietquin,
  ``Budgeted reinforcement learning in continuous state space,'' \emph{arXiv
  preprint arXiv:1903.01004}, 2019.

\bibitem{ernst2005tree}
D.~Ernst, P.~Geurts, and L.~Wehenkel, ``Tree-based batch mode reinforcement
  learning,'' \emph{Journal of Machine Learning Research}, vol.~6, pp.
  503--556, 2005.

\bibitem{leurent2019interval}
E.~Leurent, D.~Efimov, T.~Raissi, and W.~Perruquetti, ``Interval prediction for
  continuous-time systems with parametric uncertainties,'' in \emph{2019 IEEE
  58th Conference on Decision and Control (CDC)}.\hskip 1em plus 0.5em minus
  0.4em\relax IEEE, 2019, pp. 7049--7054.

\bibitem{leurent2019practical}
E.~Leurent and O.-A. Maillard, ``Practical open-loop optimistic planning,'' in
  \emph{Joint European Conference on Machine Learning and Knowledge Discovery
  in Databases}.\hskip 1em plus 0.5em minus 0.4em\relax Springer, 2019, pp.
  69--85.

\bibitem{gu2016continuous}
S.~Gu, T.~Lillicrap, I.~Sutskever, and S.~Levine, ``Continuous deep q-learning
  with model-based acceleration,'' in \emph{International Conference on Machine
  Learning}.\hskip 1em plus 0.5em minus 0.4em\relax PMLR, 2016, pp. 2829--2838.

\bibitem{wang2016dueling}
Z.~Wang, T.~Schaul, M.~Hessel, H.~Hasselt, M.~Lanctot, and N.~Freitas,
  ``Dueling network architectures for deep reinforcement learning,'' in
  \emph{International conference on machine learning}.\hskip 1em plus 0.5em
  minus 0.4em\relax PMLR, 2016, pp. 1995--2003.

\bibitem{mavrogiannis2020b}
A.~Mavrogiannis, R.~Chandra, and D.~Manocha, ``B-gap: Behavior-guided action
  prediction for autonomous navigation,'' \emph{arXiv preprint
  arXiv:2011.03748}, 2020.

\bibitem{chandra2020cmetric}
R.~Chandra, U.~Bhattacharya, T.~Mittal, A.~Bera, and D.~Manocha, ``Cmetric: A
  driving behavior measure using centrality functions,'' \emph{arXiv preprint
  arXiv:2003.04424}, 2020.

\bibitem{hicks1980causality}
J.~Hicks \emph{et~al.}, \emph{Causality in economics}.\hskip 1em plus 0.5em
  minus 0.4em\relax Australian National University Press, 1980.

\bibitem{smirnov2009granger}
D.~A. Smirnov and I.~I. Mokhov, ``From granger causality to long-term
  causality: Application to climatic data,'' \emph{Physical Review E}, vol.~80,
  no.~1, p. 016208, 2009.

\bibitem{granger1988some}
C.~W. Granger, ``Some recent development in a concept of causality,''
  \emph{Journal of econometrics}, vol.~39, no. 1-2, pp. 199--211, 1988.

\bibitem{pearl2018does}
J.~Pearl, ``Does obesity shorten life? or is it the soda? on non-manipulable
  causes,'' \emph{Journal of Causal Inference}, vol.~6, no.~2, 2018.

\bibitem{wright1934method}
S.~Wright, ``The method of path coefficients,'' \emph{The annals of
  mathematical statistics}, vol.~5, no.~3, pp. 161--215, 1934.

\bibitem{richardson2013single}
T.~S. Richardson and J.~M. Robins, ``Single world intervention graphs: a
  primer,'' in \emph{Second UAI workshop on causal structure learning,
  Bellevue, Washington}.\hskip 1em plus 0.5em minus 0.4em\relax Citeseer, 2013.

\bibitem{bareinboim2020pearl}
E.~Bareinboim, J.~Correa, D.~Ibeling, and T.~Icard, ``On pearl’s hierarchy
  and the foundations of causal inference,'' \emph{ACM Special Volume in Honor
  of Judea Pearl (provisional title)}, 2020.

\bibitem{brockman2016openai}
G.~Brockman, V.~Cheung, L.~Pettersson, J.~Schneider, J.~Schulman, J.~Tang, and
  W.~Zaremba, ``Openai gym,'' \emph{arXiv preprint arXiv:1606.01540}, 2016.

\bibitem{treiber2000congested}
M.~Treiber, A.~Hennecke, and D.~Helbing, ``Congested traffic states in
  empirical observations and microscopic simulations,'' \emph{Physical review
  E}, vol.~62, no.~2, p. 1805, 2000.

\bibitem{kesting2007general}
A.~Kesting, M.~Treiber, and D.~Helbing, ``General lane-changing model mobil for
  car-following models,'' \emph{Transportation Research Record}, vol. 1999,
  no.~1, pp. 86--94, 2007.

\bibitem{bellman2015applied}
R.~E. Bellman and S.~E. Dreyfus, \emph{Applied dynamic programming}.\hskip 1em
  plus 0.5em minus 0.4em\relax Princeton university press, 2015, vol. 2050.

\end{thebibliography}

\end{document}